%% file: main.tex
\documentclass[10pt,twocolumn,letterpaper]{article}

\usepackage{cvpr}              
\usepackage[accsupp]{axessibility}
\usepackage{arydshln}
\usepackage{booktabs}
\usepackage{mathtools}
\usepackage{algorithmicx,algorithm}
\usepackage{algpseudocode}
\usepackage{graphicx}
\usepackage{subcaption}
\input{preamble}

\definecolor{cvprblue}{rgb}{0.21,0.49,0.74}
\usepackage[pagebackref,breaklinks,colorlinks,citecolor=cvprblue]{hyperref}

\def\paperID{1806} 
\def\confName{CVPR}
\def\confYear{2024}

\title{Gradient Reweighting: Towards Imbalanced Class-Incremental Learning}

\author{Jiangpeng He\\
{\tt\small he416@purdue.edu}
\and
{Elmore Family School of Electrical and Computer Engineering, Purdue University, USA}
}

\begin{document}
\maketitle
\input{sec/0_abstract}    
\input{sec/1_intro}

\input{sec/2_related_work}
\input{sec/3_preliminary}

\input{sec/4_method_new}
\input{sec/5_experiment}

\input{sec/6_conclusion}

{
    \small
    \bibliographystyle{ieeenat_fullname}
    \bibliography{main}
}


\end{document}

%% file: preamble.tex
\usepackage[dvipsnames]{xcolor}
\newcommand{\red}[1]{{\color{red}#1}}
\newcommand{\blue}[1]{{\color{blue}#1}}


%% file: sec/0_abstract.tex
\begin{abstract}

Class-Incremental Learning (CIL) trains a model to continually recognize new classes from non-stationary data while retaining learned knowledge. A major challenge of CIL arises when applying to real-world data characterized by non-uniform distribution, which introduces a dual imbalance problem involving (i) disparities between stored exemplars of old tasks and new class data (inter-phase imbalance), and (ii) severe class imbalances within each individual task (intra-phase imbalance). We show that this dual imbalance issue causes skewed gradient updates with biased weights in FC layers, thus inducing over/under-fitting and catastrophic forgetting in CIL. Our method addresses it by reweighting the gradients towards balanced optimization and unbiased classifier learning. Additionally, we observe imbalanced forgetting where paradoxically the instance-rich classes suffer higher performance degradation during CIL due to a larger amount of training data becoming unavailable in subsequent learning phases. To tackle this, we further introduce a distribution-aware knowledge distillation loss to mitigate forgetting by aligning output logits proportionally with the distribution of lost training data. We validate our method on CIFAR-100, ImageNetSubset, and Food101 across various evaluation protocols and demonstrate consistent improvements compared to existing works, showing great potential to apply CIL in real-world scenarios with enhanced robustness and effectiveness. 

\end{abstract}

%% file: sec/1_intro.tex
\section{Introduction}
\label{sec:intro}
The ever-evolving and unpredictable nature of real-world environments drives the imperative to develop Class-Incremental Learning (CIL) systems with the capability of acquiring knowledge continuously from non-stationary data where new classes appear sequentially over time. The advantage of CIL resides in both memory and computational efficiency which eliminates the requirement of storing all previously learned data or retraining the model from scratch entirely, making CIL applicable to various practical applications such as on-device learning~\cite{Hayes_2022_Embedded_CL}. However, deep neural network suffers from catastrophic forgetting~\cite{CF} when learning new classes, where the performance on learned classes decreases significantly due to the unavailability of old training data. Moreover, conventional CIL methodologies typically tackle this challenge by assuming a balanced data distribution, where each class has roughly the same number of training samples. Nevertheless, this assumption is misaligned with real-world scenarios where data is usually long-tail distributed with significant disparity between instance-rich (\textit{head}) classes and instance-rare (\textit{tail}) classes. Such imbalance presents unique challenges that conventional CIL methods cannot adequately address, thus reducing their effectiveness and broader applicability. 

\begin{figure}[t]
\begin{center}
  \includegraphics[width=1.\linewidth]{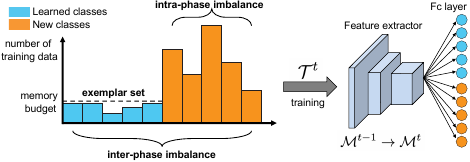}
  \vspace{-2ex}\caption{The illustration of imbalanced class-incremental with a dual imbalance issue including the intra-phase imbalance within each new task $\mathcal{T}$ and inter-phase imbalance between old tasks exemplars and new task training data. $\mathcal{M}^{t}$ refers to the model after learning the new task $\mathcal{T}^t$.}
  \label{fig:introduction}
  \vspace{-0.8cm}
\end{center}
\end{figure}

As shown in Figure~\ref{fig:introduction}, CIL with imbalanced data introduces a dual challenge encompassing both inter-phase and intra-phase imbalances. Conventional CIL approaches primarily address the inter-phase imbalance while the intra-phase imbalance emerges distinctly in the context of imbalanced CIL. In general, the inter-phase imbalance arises from the disparities between new class data and the learned task data preserved in the exemplar set for knowledge replay~\cite{ICARL, GEM}. Such imbalances can skew the learning process of the classifier between new and old classes\cite{guo2023dealing, MAFDRC} (\textit{e.g.}, the learned weights in the FC layer are heavily biased~\cite{BiC, mainatining}), leading to biased predictions towards the newer classes and thus inducing catastrophic forgetting. In the case of intra-phase imbalance, either the new classes or the stored exemplars individually present their own imbalanced distributions (the imbalance of the exemplar set arises when the number of training data is less than the memory budget for each class). This not only affects the learning of new knowledge but also exacerbates the forgetting issue by intensifying the inter-phase biases. 

One of the major challenges posed by learning from imbalanced data is the biased gradient updates in FC layers that are dominated by instance-rich classes. Specifically, in gradient-based optimization such as stochastic gradient descent (SGD), the weight update step in each iteration is heavily influenced by the magnitude of gradients, which significantly depends on the data distribution of the sampled mini-batch. As a result, instance-rich classes tend to receive gradients with larger magnitudes thus skewing the optimization process to make the classifier over-fit on head classes while under-fit on tail classes. This problem becomes more challenging under the context of CIL as the training data distribution changes over time, and the learned classes suffer from catastrophic forgetting. Besides, we observe that the forgetting could also be imbalanced where the head classes usually suffer more performance degradation as a substantial portion of their training data becomes unavailable in subsequent incremental learning stages. To this end, our work aims to address the dual imbalance issue by reweighting the gradient updates in the FC layer, which recalibrates the optimization process and fosters the learning of unbiased classifiers. Furthermore, we introduce a distribution-aware knowledge distillation loss to alleviate the imbalanced catastrophic forgetting problem by considering the distribution of lost training data to impose stronger regularization effects on the head classes. The main contributions of this work are summarized in the following,
\begin{itemize}
    \item We study CIL under realistic class-imbalanced scenarios and introduce a new end-to-end gradient reweighting framework to tackle both intra-phase and inter-phase imbalance challenges by rebalancing the optimization process in FC layers.

    \item A new distribution-aware knowledge distillation loss is proposed to further mitigate imbalance catastrophic forgetting caused by the uneven number of lost training data during imbalanced CIL.
 
    \item The efficacy of our proposed method is validated through extensive experiments across a variety of CIL settings, which achieves notable improvements over existing methods in both CIL and long-tailed recognition tasks. 
\end{itemize}



%% file: sec/2_related_work.tex
\vspace{-0.2cm}
\section{Related Work}
\vspace{-0.2cm}
\label{sec:related_work}
In this section, we review the existing studies that are most related to our work including class-incremental learning in Section~\ref{subsec: relatedwork_cil} and long-tailed recognition in Section~\ref{subsec: relatedwork_lt}. 
\subsection{Class-Incremental Learning}
\vspace{-0.1cm}
\label{subsec: relatedwork_cil}
Class-incremental learning (CIL) is a subarea of continual learning where the task index is not provided during the inference phase, \textit{i.e. }a single head classifier~\cite{SIT} is used for all classes seen so far. Existing CIL has been studied in both online and offline scenarios where each data is observed only once by the model in the former case. 

\textbf{Conventional CIL} approaches can be mainly categorized into three groups including regularization, memory replay, and parameter isolation. 
The \textit{regularization-based} methods aim to restrict drastic alterations to model parameters that are important for learned classes. A representative technique is to apply knowledge distillation~\cite{KD} based on output logits~\cite{LWF, ICARL, EEIL, BiC} or the intermediate layers~\cite{rebalancing, douillard2020podnet, simon2021learning_featkd}. In addition, recent studies~\cite{BiC, SSIL, dualmemory} reveal the biased weights in FC layer towards newer learned classes and address it by applying post-hoc biased logits correction~\cite{BiC, mainatining, dualmemory, He_2022_WACV} or using separated softmax~\cite{SSIL} and cosine normalized classifier~\cite{rebalancing}.
The \textit{replay-based} methods store a small set of exemplar data to perform knowledge rehearsal in subsequent phases. Herding algorithm~\cite{HERDING} is widely employed in CIL~\cite{ICARL, EEIL, BiC} for exemplar selection based on the class mean. Later, a learning-based exemplar selection method is introduced in~\cite{liu2020mnemonics}. In the online setting, the exemplar selection is performed at the end of each training iteration~\cite{GEM, prabhu2020gdumb_online, NEURIPS2019_e562cd9c_GSS, NEURIPS2019_15825aee_MIR, guo2023dealing} without knowing the distribution of training data. 
Finally, the \textit{parameter isolation} related approaches gradually expand the network size or increase model parameters~\cite{DER, foster, liu2021adaptive, DEN, zenke2017continual_splitcifar} to provide dedicated space for learning new knowledge while ensuring the previously learned information remains undisturbed. 

\textbf{Imbalanced CIL} remains less explored compared to conventional CIL due to the intricate challenge of addressing both catastrophic forgetting and imbalanced learning simultaneously. The earlier works focused on imbalanced CIL under multi-label~\cite{kim2020imbalanced_PRS}, semi-supervised~\cite{belouadah2020active}, and online scenarios~\cite{chrysakis2020online}. The most recent study~\cite{liu2022long} formulated long-tailed CIL for both shuffled and ordered cases and proposed a two-stage technique to decouple representation learning and classifier learning. Later, a dynamic residual classifier is introduced in~\cite{MAFDRC} to handle the imbalance between new and learned classes. In this work, we primarily focus on CIL following the shuffled long-tailed case~\cite{liu2022long} to address the imbalance issue from the perspective of balancing the gradient updates in FC layer. In addition, contrasting the findings in~\cite{kim2020imbalanced_PRS}, we observed imbalanced forgetting where the head classes instead suffer more performance degradation attributed to the larger volume of lost training data in subsequent incremental phases. Therefore, we incorporate a distribution-aware knowledge distillation loss to impose a more stringent regularization on head classes to further mitigate the forgetting issue. 

\subsection{Long-tailed Learning}
\label{subsec: relatedwork_lt}
The deep long-tailed learning for visual recognition has been studied comprehensively~\cite{zhang2021deep, yang2022survey} over the decades due to the prevalence of class-imbalanced data obtained in real-world scenarios. The major challenge arises from the fact that the performance of deep learning models tends to be dominated by head classes, leaving the learning for tail classes significantly under-realized. In this work, we center on class-rebalanced approaches with the goal of re-balancing the influence resulting from imbalanced training samples, which mainly consists of re-sampling, and class-sensitive learning. Specifically, the \textit{re-sampling} based methods~\cite{ROS, RUS_ROS, CMO} aim to construct a balanced training batch by increasing the probability of tail classes to be sampled. The \textit{class-sensitive} learning seeks to adjust the value of training loss to rebalance the uneven learning effects by reweighting class influence~\cite{cui2019class, BSLoss, LDAM, IBLoss, logit_adjust}. In addition, recent studies~\cite{tan2020equalizationv1, tan2021equalizationv2, guo2023dealing} argue that the imbalance between positive and negative gradients severely inhibits balanced learning and they address it by adjusting the loss influence across different classes. Our proposed method stems from a similar observation but identifying the magnitude of gradient vectors diverges significantly between head and tail classes during the learning process. This discrepancy leads to uneven optimization steps, subsequently causing the model to over-fit the head classes while under-fitting the tail classes. To address this, our goal is to equilibrate the gradient updates, ensuring a more balanced learning process. Furthermore, existing long-tailed learning methods do not consider previously acquired knowledge for learned classes, making them unsuitable under the context of CIL. 

%% file: sec/3_preliminary.tex
\vspace{-1ex}
\section{Preliminaries}
\vspace{-1ex}
\label{sec:prelim}
The CIL can be formulated as learning a sequence of $N$ tasks $\{\mathcal{T}^1, \mathcal{T}^2,...\mathcal{T}^N \}$. Each task $t \in [1, N]$ refers to one learning phase in CIL, which can be represented as $\mathcal{T}^t = \{\mathcal{X}^t, \mathcal{Y}^t\}$ where $\mathcal{X}^t = \{\textbf{x}^t_1, ..., \textbf{x}_{|\mathcal{X}^t|}^t\}$ denotes the set of total $|\mathcal{X}^t|$ number of training data and $\mathcal{Y}^t$ denotes the associated class labels belonging to $|\mathcal{Y}^t|$ classes. In general, we have total $\sum_{t=1}^N |\mathcal{X}^t|$ training data corresponding to $\sum_{t=1}^N|\mathcal{Y}^t|$ classes. After learning each task $\mathcal{T}^{t}$, the model is evaluated on test data belonging to all seen classes $\mathcal{Y}^{1:t}$ without knowing the task identifier $t$. A strong constraint in CIL when incrementally learning a new task $\mathcal{T}^{t} (t > 1)$ is the unavailability of previously learned tasks data $\{\mathcal{X}^1,...,\mathcal{X}^{t-1}\}$, which causes the catastrophic forgetting on learned tasks $\mathcal{T}^{1:t-1}$. Therefore, given a memory budget $n_\varepsilon$ data per class, the replay-based methods employ an exemplar set $\mathcal{E}^t = \{\mathcal{X}^{1:t-1}_e, \mathcal{Y}^{1:t-1}_e \}$ to store both data $\mathcal{X}^{1:t-1}_e$ and their class labels $\mathcal{Y}^{1:t-1}_e$ to combine with new data $\{\mathcal{X}^{1:t-1}_e \cup \mathcal{X}^{t}\} \times \{\mathcal{Y}^{1:t-1}_e \cup \mathcal{Y}^{t}\} $ during each incremental learning phase to maintain the learned knowledge.

\textbf{Imbalanced CIL} refers to the case where the number of training data varies a lot among different classes while the test data remain balanced. This induces a dual imbalance issue in the training phase including the intra-phase imbalance and inter-phase imbalance. Specifically, the former case represents the discrepancy within each task $\mathcal{T}^t$ where given a head class $j$ and a tail class $k$ with $j, k \in \mathcal{Y}^{t}$, we have the corresponding training data $ n_{j} \gg n_{k}$. Furthermore, the exemplar set $|\mathcal{E}^t|$ may also exhibit this class-imbalance issue especially given a larger memory budget $n_{\varepsilon}$ as more training from head classes will be retained compared to tail classes, thereby amplifying this challenge. The latter case refers to the discrepancy between new task data and stored exemplars when $t > 1$ where $|\mathcal{X}^t| \gg |\mathcal{X}_e^{1:t-1}|$. In general, the intra-phase imbalance can restrict the model's capacity to learn new knowledge while the inter-phase imbalance can lead to catastrophic forgetting as the model might exhibit a bias towards new classes.

\begin{figure}[t]
\begin{center}
  \includegraphics[width=1.\linewidth]{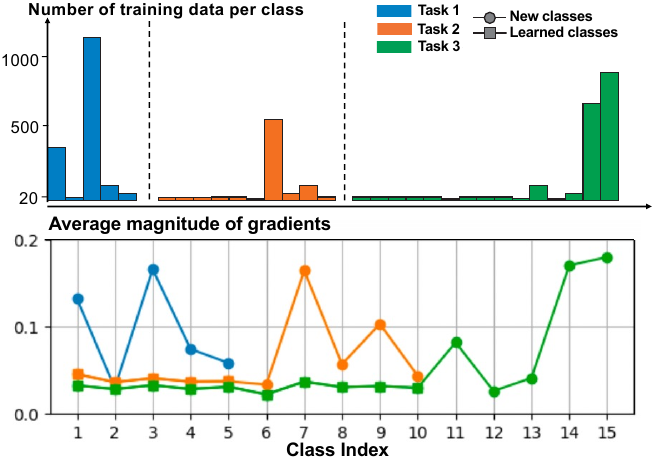}
  \vspace{-5ex}\caption{The average magnitudes of gradient $||\nabla_{\mathcal{L}_{ce}}(W^j)||$ for each class $j$ by incrementally learning 3 tasks $\mathcal{T}^1,\mathcal{T}^2,\mathcal{T}^3$ with cross-entropy and memory budget $n_{\varepsilon} = 20$ exemplars per class. }
  \label{fig:gradient_norm}
  \vspace{-0.5cm}
\end{center}
\end{figure}

\begin{figure*}[t]
\begin{center}
  \includegraphics[width=1.\linewidth]{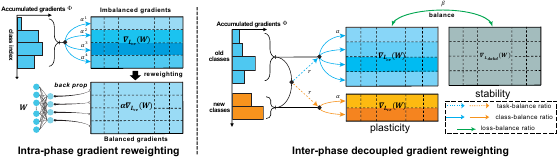}
  \vspace{-4ex}\caption{The overview of gradient reweighting under imbalanced CIL. Given the classifier $W$, the intra-phase gradient weighting is guided by scaling the gradient matrix $\nabla_{\mathcal{L}_{ce}}(W)$ with class balance ratios $\alpha$ derived from the cumulative gradients $\Phi$ over iterations. Concurrently, the inter-phase Decomposed Gradient Reweighting (DGR) balances the plasticity learning by separately adjusting gradients with class-balance ratios $\alpha$ and task-balance ratios $r$. Followed by tuning the stability-plasticity trade-off with a loss balance ratio $\beta$.}
  \label{fig:method}
  \vspace{-0.4cm}
\end{center}
\end{figure*}

\subsection{Imbalanced Gradients}
\label{subsec: imbalanced gradients}
Imbalanced gradient updates pose a significant challenge when learning from imbalanced data as studied in~\cite{francazi2023theoretical}. Generally, consider the weights of a classifier $W \in \mathbb{R}^{d_f \times c}$ with input feature dimension $d_f$ and output classes $c$. Given a loss function $\mathcal{L}$, the gradient-based weight update process is formulated as 
\begin{equation}
\label{eq: gradeint_update}
    \begin{aligned}
        W_{i+1} = W_i - \eta \nabla_{\mathcal{L}}(W_i)
    \end{aligned}
\end{equation}
where $i$ denotes the index of iteration and $\eta$ is the learning rate. The gradient matrix $\nabla_{\mathcal{L}}(W_i) \in \mathbb{R}^{d_f \times c}$ determines the magnitude of the update $\eta ||\nabla_{\mathcal{L}}(W_i)|| $ and the direction of the update $\frac{\nabla_{\mathcal{L}}(W_i)}{||\nabla_{\mathcal{L}}(W_i)||}$ where $||\cdot||$ denotes the $l_2$ norm. In general, the magnitude of the gradient is proportional $||\nabla_{\mathcal{L}}(W_i)|| \propto \mathcal{L}$ to the loss value which is usually computed as the average loss over all data samples. However, challenges arise when the gradient is estimated from a mini-batch containing class-imbalanced data, \textit{e.g.}, $n_{j} \gg n_{k}$ given a head class $j$ and tail class $k$. In this case, the gradient magnitude for the head class tends to be significantly larger than that for the tail class $||\nabla_{\mathcal{L}}(W_i^{j})|| \gg ||\nabla_{\mathcal{L}}(W_i^{k})||$.
In Figure~\ref{fig:gradient_norm}, we highlight the gradient imbalances across different classes by calculating the average gradient magnitude at the end of each task by incrementally learning 3 new tasks (with cross-entropy loss) on ImageNetSubset-LT~\cite{liu2022long} with each task contains 5 classes. This disparity in gradient magnitudes leads to an imbalanced optimization process where larger gradients from head classes cause larger optimization steps in weight updates, thus potentially leading to over-fitting. Conversely, smaller updates for tail classes may result in under-fitting. Furthermore, the larger gradient magnitudes can also induce biases in the norm of weight vectors as $||W_{i+1} - W_i|| \propto ||\nabla_{\mathcal{L}}(W_i)||$, resulting in biased prediction towards instance-rich or newly learned classes under CIL. In the following section, we illustrate our proposed method to address the aforementioned issues by re-weighting the biased gradients to facilitate learning of a balanced classifier.

%% file: sec/4_method_new.tex
\section{Method}
\label{sec:method}
\textbf{Overview.} As illustrated in Figure~\ref{fig:method}, we aim to reweight the imbalanced gradient matrix by classwisely multiplying it with a balance vector $\boldsymbol{\alpha} = [\alpha^1, \alpha^2,...\alpha^c]$ for total $c$ classes seen so far. Therefore, the gradient update process can be modified as \vspace{-1ex}
\begin{equation}
\vspace{-1ex}
\label{eq: gradeint_update}
        W_{i+1}^j = W_i^j - \eta \alpha^j \nabla_{\mathcal{L}}(W_i^j)
\end{equation}
where $W_i^j$ refers to the weight vector in FC layer for class $j$ and $\alpha^j \nabla_{\mathcal{L}}(W_i^j)$ denotes the corresponding reweighted balanced gradient. However, it is non-trivial to obtain the appropriate balance vector due to the intricate data-dependent optimization process. In addition, when learning a new task $\mathcal{T}^t(t>1)$ with the present of learned classes belonging to $\mathcal{Y}^{1:t-1}$, it poses two additional challenges. First, the training distribution, denoted as $\mathcal{D}$, changed for learned classes during CIL where $\mathcal{D}(\mathcal{X}_e^{1:t-1}, \mathcal{Y}_e^{1:t-1}) \neq \mathcal{D}(\mathcal{X}^{1:t-1}, \mathcal{Y}^{1:t-1})$ as only limited exemplars are stored for training in subsequent learning phases after they were initially observed. Therefore, simply balancing all gradients without accounting for this will result in knowledge forgetting for learned classes, which could be imbalanced as well due to the uneven number of lost training data between head and tail classes. Second, the training objective varies between new and old tasks where the learned tasks $\mathcal{T}^{1:t-1}$ build upon knowledge accumulated from previous learning phases, while the new task $\mathcal{T}^{t}$ starts from scratch without any prior knowledge. Consequently, balancing the gradient updates for all classes may not be sufficient due to the intrinsic imbalance in optimization, which requires a more adaptive and flexible approach to ensure the model can effectively learn new tasks without forgetting the knowledge from previous ones.

In the following sections, we illustrate how to effectively determine the balance vector $\boldsymbol{\alpha}$ under CIL for intra-phase scenario in Section~\ref{subsec: intra-phase}, and inter-phase case to tackle both biased gradients and imbalanced forgetting in Section~\ref{subsec: inter-phase}.

\subsection{Intra-Phase Class-Imbalance}
\label{subsec: intra-phase}
Consider the case of learning the first task $\mathcal{T}^1 = \{\mathcal{X}^1, \mathcal{Y}^1\}$ from a scratch model $\mathcal{M}^0(f_\theta, W)$ where $f_\theta$ and $W$ refers to the feature extractor and classifier, respectively. The cross-entropy loss is formulated as 
\begin{equation}
\label{eq: cn}
    \begin{aligned}
        \mathcal{L}_{ce} = - \frac{1}{|\mathcal{X}^1|}\sum_{k=1}^{|\mathcal{X}^1|} \sum_{j=1}^{|\mathcal{Y}^1|}\hat{y}_j \log(p_j)
    \end{aligned}
\end{equation}
where $p_j$ is the Softmax of the $j$th output logit $z_j = [W^Tf_\theta(\textbf{x}_k)]_j$ and $\hat{y}_j$ denotes the class label. The goal is to determine the class-wise balance vector $\boldsymbol{\alpha} = [\alpha^1, ...\alpha^{|\mathcal{Y}^1|}]$. Recognizing that a static balance vector may not sufficiently capture the biases inherent in the gradients matrix that evolve as the learning progresses, we propose a simple yet effective solution in this work to determine the balance vector adaptively by leveraging historical accumulated gradients. In detail, we calculate the class balance ratio $\alpha_i^j$ at iteration $i$ based on the accumulated magnitude of gradients corresponding to each class $j$ as \vspace{-1ex}
\begin{equation}
\vspace{-1ex}
\label{eq: alpha-intra-phase}
    \begin{aligned}
    \alpha^{j}_i = \frac{\underset{m \in \mathcal{Y}^1}{\textit{min}}\Phi^m_i}{\Phi^{j}_i}, \quad 
        \Phi^{j}_i = \sum_{n=1}^i ||\nabla_{\mathcal{L}_{ce}}(W_{n}^{j})||
    \end{aligned}
\end{equation}
By doing so, we dynamically recalibrate the weight updates, ensuring that both majority and minority classes contribute equally to the learning process and address the biased issue. 

\textbf{Regularized softmax cross-entropy.} While the gradient update adjustments help in emphasizing tail classes, it also inadvertently leads to increased gradients towards head classes due to the decrease of output logit $z_j$ of head classes, making the cross-entropy loss put more effort into increasing its output logits. Motivated by~\cite{logit_adjust}, we alleviate this side effect by applying regularized cross-entropy using a modified softmax equation 
\begin{equation}
\label{eq:r-softmax}
    \begin{aligned}
    p_j = \frac{exp(z_j + \log \pi_j)}{\sum_{m = 1}^{|\mathcal{Y}^1|}exp(z_m + \log \pi_m)}
    \end{aligned}
\end{equation}
where $\pi_j$ denotes the estimates of class priors and $\log \pi_j$ is the per-label offsets added to the original output logit. In practice, the calibrated class distribution with the number of data for each seen class is widely used to determine $\pi_j$. Generally, the instance-rich classes have larger $\pi_j$ with an increase of softmax output $p_j$ to compensate for the side effect of down-weighting the gradients during the training process. 


\subsection{Inter-Phase Class-Imbalance}
\label{subsec: inter-phase}
We extend our intra-phase method to address the inter-phase imbalance in CIL by adding two modules: (i) a Distribution-Aware Knowledge Distillation (DAKD) loss to maintain the information from the original training distribution while mitigating the imbalanced forgetting (Section~\ref{subsubsec: im-forgetting}), and (ii) a Decoupled Gradient Reweighting (DGR) framework to separately manage the gradients between new and already learned tasks to ensure a balance between the stability of past knowledge and the plasticity required for learning new classes (Section~\ref{subsubsec: integrated gradient}). 

\vspace{-1ex}
\subsubsection{Distribution-Aware Knowledge Distillation}

\label{subsubsec: im-forgetting}
Similar to existing bias correction methods~\cite{BiC, mainatining, SSIL}, we consider adding the knowledge distillation loss~\cite{KD} to maintain the learned knowledge due to the changing of training distribution for learned classes. The original distillation loss for learning a new task $t$ can be expressed as 
\begin{equation}
\label{eq: kd}
    \begin{aligned}
        \mathcal{L}_{kd}(z, \hat{z}) = - \frac{1}{|\mathcal{X}^{t}+\mathcal{X}_e^{1:t-1}|}\sum_{k=1}^{|\mathcal{X}^{t}+\mathcal{X}_e^{1:t-1}|} \sum_{j=1}^{|\mathcal{Y}^{1:t-1}|}\hat{p}_j^\tau \log(p_j^\tau)
    \end{aligned}
\end{equation}
where $\hat{p}_j^\tau = \textit{Softmax} (\hat{z}_j/\tau)$ is softened ($\tau = 2$) output probability from the teacher model $\hat{M}^{t-1}$ with frozen parameters obtained in the last learning phase $t-1$.
However, as mentioned earlier in Section~\ref{sec:intro}, catastrophic forgetting can be imbalanced as a larger volume of instances from the head class become unavailable in comparison to the tail class in the subsequent incremental learning phases due to the fixed memory budget. Though the most intuitive way to address this issue is storing more data for head classes and less data for tail classes, it poses another nontrivial challenge arising from a class-imbalanced exemplar set that potentially intensifies both over-fitting and under-fitting issues. Therefore, we introduce the Distribution-Aware Knowledge Distillation (DAKD) to maintain the knowledge by taking into account the distribution of the lost training data. Specifically, we obtain the lost training data distribution $\mathbf{s}$ by calculating $s_j = |\mathcal{X}^j| - |\mathcal{X}_e^{j}|$ for each class $j \in \mathcal{Y}^{1:t-1}$ where $|\mathcal{X}^j|$ and $|\mathcal{X}_e^{j}|$ denote the number of original training data and stored exemplars for class $j$, respectively. Motivated by~\cite{DCKD}, we decouple the original distillation loss into a weighted sum of two parts using a ratio $\sigma \in [0, 1]$ (1 indicates balanced distribution) measured by the entropy of $\mathbf{s}$. The DAKD is then formulated as 
\begin{equation}
\label{eq: decompkd}
    \begin{aligned}
        \mathcal{L}_{dakd}(z, \hat{z}|\mathbf{s}) = \sigma \mathcal{L}_{kd} (z,\hat{z})  + (1 - \sigma) \mathcal{L}_{kd}^{imb}(\tilde{z},\hat{z}) 
    \end{aligned}
\end{equation}
where $\mathcal{L}_{kd}$ denotes the balanced part calculated using the original output logit $z$, and $\mathcal{L}_{kd}^{imb}$ denotes the imbalanced part with adjusted output logits $\tilde{z}$ determined by
\begin{equation}
\label{eq: z}
    \begin{aligned}
        \tilde{z}_j = \frac{s_j}{\sum_m^{|\mathcal{Y}^{1:t-1}|} s_m} z_j + (1 - \frac{s_j}{\sum_m^{|\mathcal{Y}^{1:t-1}|} s_m})\hat{z}_j 
    \end{aligned}
\end{equation}
where this calibrated distribution of output logits $\tilde{z}$ demonstrates larger discrepancy ($|\tilde{z}_j - \hat{z}_j|$) for class $j$ with a higher volume of lost training data (\textit{i.e.}, head classes), thereby assigning those classes with more efforts for knowledge distillation compared to those with less data lost (\textit{i.e.}, tail classes) which only require a subtler distillation intervention to maintain the learned knowledge. This tailored logit adjustment ensures that the extent of knowledge distillation is appropriately aligned with the level of data lost experienced by each class. 
The overall loss function for learning a new task $\mathcal{T}^t, t>1$ can be expressed as 
\begin{equation}
\label{eq: z}
    \begin{aligned}
        \mathcal{L} = \mathcal{L}_{ce} + \lambda \mathcal{L}_{dakd}
    \end{aligned}
\end{equation}
We use $\lambda = \lambda_b \sqrt{|\mathcal{X}_{old}|/|\mathcal{X}_{new}|}$ to adjust the influence of knowledge distillation, which increases as more data have been observed. $|\mathcal{X}_{old}|$ and $|\mathcal{X}_{new}|$ denote the number of old and new classes training data and $\lambda_b$ is a fixed scalar.

\subsubsection{Decoupled Gradient Reweighting}
\label{subsubsec: integrated gradient}
The overview of Decoupled Gradient Reweighting (DGR) is shown in Figure~\ref{fig:method}, which addresses the inter-phase imbalance issue by striking a balance between stability (maintain past knowledge) and plasticity (learn new classes). Generally, the gradients from cross-entropy loss ($\nabla_{\mathcal{L}_{ce}}$) in CIL represent the plasticity that enables the model to adapt to new training data distribution by incorporating both exemplars of old classes and data from new classes. In contrast, the stability is introduced by the gradients from knowledge distillation loss ($\nabla_{\mathcal{L}_{dakd}}$) to guide the model towards a solution that aligns with the training distribution from the previous learning phases. Therefore, the DGR address inter-phase imbalance in two folds by first reweighting gradient $\nabla_{\mathcal{L}{ce}}$ from $\mathcal{L}_{ce}$ to ensure unbiased plasticity and then modulating the interaction between $\nabla{\mathcal{L}_{ce}}$ and $\nabla{\mathcal{L}_{dakd}}$ to attain a balanced equilibrium between plasticity and stability.

\noindent \textbf{For plasticity}, the DGR first separately reweights the gradients $\nabla_{\mathcal{L}_{ce}}$ for learned classes $j \in \mathcal{Y}^{1:t-1}$ and new classes for $j \in \mathcal{Y}^{t}$ by calculating class-balance ratios $\alpha_i^j$  respectively as specified in Equation~\ref{eq: alpha-intra-phase}. Following this, a task-balance ratio $r_i^j$ is introduced for tuning between new and learned tasks as
\begin{equation}
\label{eq:ri}
    \begin{aligned} 
    r_i^j = \left\{ \begin{array}{ccl}
   \textit{min} \{1, \frac{1}{r_{\Phi_i}}\} & j \in \mathcal{Y}^{1:t-1} \\  \textit{min} \{1, r_{\Phi_i} \times exp( - \gamma \frac{|\mathcal{X}^{1:t-1}|}{|\mathcal{X}^{1:t}|}) \} & j \in \mathcal{Y}^{t}
    \end{array}\right.
    \end{aligned}
\end{equation}
where $exp( - \gamma \frac{|\mathcal{X}^{1:t-1}|}{|\mathcal{X}^{1:t}|})$ is the attenuation factor and $\gamma >0 $ is a hyper-parameter to adjust its magnitude. $r_{\Phi_i} = \frac{\overline{\Phi}_i^{j \in \mathcal{Y}^{1:t-1}}}{\overline{\Phi}_i^{j \in \mathcal{Y}^{t}}} $ is the ratio of mean accumulated gradients for learned classes $\overline{\Phi}_i^{j \in \mathcal{Y}^{1:t-1}}$ to new classes $\overline{\Phi}_i^{j \in \mathcal{Y}^{t}}$. 

The advantage of DGR with an attenuation factor lies in promoting a more equitable optimization based on the fact that the learned classes are built upon the knowledge learned in prior learning phases, whereas the new classes are being trained from scratch. Therefore, merely balancing the gradient contributions between new and old classes could inadvertently lead to the under-fitting of new classes as they may not receive adequate training emphasis. Recognizing this, the attenuation factor adjusts the gradient reweighting ratio in favor of new classes proportional to the volume of data that has already been observed so far $|\mathcal{X}^{1:t-1}|$, thereby providing a calibrated boost in support of the new classes to mitigate the learning disparities. 

\noindent \textbf{For tuning between stability and plasticity}, we adjust the magnitude of gradients from distillation $||\nabla_{\mathcal{L}_{dakd}}(W_i)||$ to make it balance with reweighted cross-entropy by including a loss balance ratio $\beta_i$ at iteration $i$ as 

\begin{equation}
\label{eq:betai}
    \begin{aligned} 
    \beta_i = \frac{||\alpha_i r_i \nabla_{\mathcal{L}_{ce}}(W_i)||}{||\nabla_{\mathcal{L}_{dakd}}(W_i)||}
    \end{aligned}
\end{equation}

The overall DGR can be formulated as 
\begin{equation}
\label{eq: gradient-inter-phase}
    \begin{aligned}
        W_{i+1}^{j} = W_i^{j} - \mathrlap{\underbrace{\phantom{\quad \quad \quad (\alpha^{j}_i \nabla_{\mathcal{L}_{ce}}}}_{j \in \mathcal{Y}^{t}}} \mathrlap{\overbrace{\phantom{ \nabla_{\mathcal{L}_{ce}}(W_i^{j}) + \beta_{i} \nabla + \beta_{i} \quad \quad \quad \quad }}^{j \in \mathcal{Y}^{1:t-1}}} \eta (
        \alpha^{j}_i r_i^j \nabla_{\mathcal{L}_{ce}}(W_i^{j}) + \beta_{i} \nabla_{\mathcal{L}_{dakd}}(W_i^{j}))
    \end{aligned}
\end{equation}
Note that in this work, we intentionally avoid classwisely reweighting the gradients from knowledge distillation loss as it contains instrumental information obtained from previous training distribution to help maintain the discrimination of learned classes during CIL as studied in~\cite{mainatining}. Later in Section~\ref{subsec: ablation}, we will show that reweighting the gradients of knowledge distillation could harm the overall performance. 

%% file: sec/5_experiment.tex
\begin{figure*}[t]
\begin{center}
  \includegraphics[width=1.\linewidth]{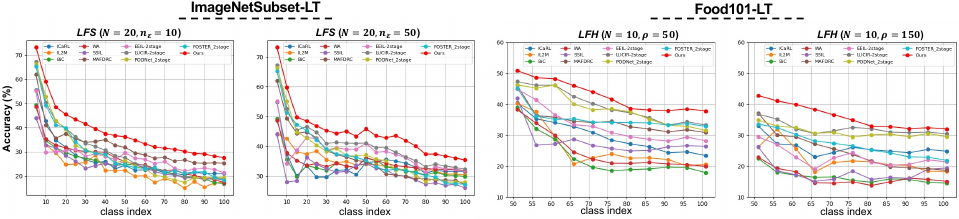}
  \vspace{-5ex}\caption{The classification accuracy (\%) on test data belonging to all classes seen so far at each incremental step by varying the memory budget $n_\varepsilon \in \{10, 50\}$ on ImageNetSubset-LT and imbalance factor $\rho \in \{50, 150\}$ on Food101-LT. }
  \label{fig:results_all}
  \vspace{-0.5cm}
\end{center}
\end{figure*}
\begin{table*}[t!]
    \centering
    \scalebox{.93}{
    \begin{tabular}{lcccccccccccc}
        \hline
          \multicolumn{1}{c}{Datasets} &\multicolumn{4}{c}{\textbf{CIFAR100-LT}} & \multicolumn{4}{c}{\textbf{ ImageNetSubset-LT}} & \multicolumn{4}{c}{\textbf{Food101-LT}} \\

          \multicolumn{1}{c}{Evaluation protocol}  & \multicolumn{2}{c}{\textit{LFS}} & \multicolumn{2}{c}{\textit{LFH}} & \multicolumn{2}{c}{\textit{LFS}} & \multicolumn{2}{c}{\textit{LFH}} & \multicolumn{2}{c}{\textit{LFS}} & \multicolumn{2}{c}{\textit{LFH}}\\
        \cdashline{2-13}
          \multicolumn{1}{c}{Total tasks $N$}  & 10 & 20 & 5 & 10 & 10 & 20 & 5 & 10 & 10 & 20 & 5 & 10\\
         \hline
        iCaRL~\cite{ICARL}& 21.83 & 24.28 & 28.68 & 28.33 & 33.75 & 29.71 & 41.82 & 40.21 & 18.13 & 12.50 & 21.83 & 21.31 \\
        IL2M~\cite{dualmemory}& 31.37 & 29.99 & 34.90 & 33.42 & 31.70 & 25.20 & 40.75 & 39.08 & 16.11 & 16.27 & 23.93 & 22.48 \\
        BiC~\cite{BiC} & 28.89 & 20.10 & 25.68 & 25.95 & 33.31 & 30.86 & 33.18 & 29.23 & 16.94 & 16.81 & 22.80 & 20.75 \\
        WA~\cite{mainatining} & 27.63 & 23.48 & 32.07 & 26.85 & 32.58 & 29.03 & 32.62 & 28.10 & 16.58 & 15.99 & 18.45 & 19.45 \\
        SSIL~\cite{SSIL} & 26.07 & 26.15 & 30.72 & 29.21 & 30.38 & 25.99 & 38.97 & 35.18 & 16.86 & 15.65 & 21.65 & 19.03 \\
        FOSTER~\cite{foster} & 30.43 & 29.96 & 37.25 & 37.91 & 34.38 & 29.75 & 46.51 & 43.88 & 24.27 & 20.45 & 32.39 & 31.46 \\
        MAFDRC~\cite{MAFDRC} & 32.67 & 31.95 & 37.94 & 38.51 & \blue{\textbf{40.01}} & 34.48 & 48.23 & 44.12 & 26.93 & 19.21 & 34.22 & 30.91 \\
        EEIL-2stage~\cite{EEIL,liu2022long} & \blue{\textbf{33.64}} & \blue{\textbf{32.25}} & 36.40 & 34.91 & 36.84 & 30.39 & 43.62 & 41.49 & 19.75 & 20.02 & 22.65 & 22.83 \\
        LUCIR-2stage~\cite{rebalancing,liu2022long} & 31.09 & 31.03 & 38.47 & 37.86 & 39.87 & \blue{\textbf{34.79}} & \blue{\textbf{48.97}} & 47.39 & \blue{\textbf{27.65}} & \blue{\textbf{24.68}} & \blue{\textbf{36.05}} & 35.06 \\
        PODNet-2stage~\cite{douillard2020podnet,liu2022long} & 30.41 & 30.37 & 38.38 & 38.45 & 35.47 & 31.71 & 48.02 & \blue{\textbf{47.74}} & 23.78 & 21.13 & 35.42 & \blue{\textbf{35.22}} \\
        FOSTER-2stage~\cite{foster,liu2022long} & 31.27 & 30.68 & \red{\textbf{40.26}} & \red{\textbf{39.43}} & 36.47 & 33.95 & 48.89 & 46.93 & 25.82 & 22.28 & 35.69 & 33.48 \\
          \hline
        Ours & \textbf{\red{35.66}} & \red{\textbf{34.35}} & \blue{\textbf{40.18}} & \blue{\textbf{39.11}} & \red{\textbf{45.12}} & \red{\textbf{40.79}} & \red{\textbf{50.57}} & \red{\textbf{49.13}} & \red{\textbf{29.05}} & \red{\textbf{26.42}} & \red{\textbf{36.84}} & \red{\textbf{36.19}} \\
        \hline
    \end{tabular}
    }
    \vspace{-2ex}
        \caption{Results of average accuracy (\%) on CIFAR100-LT, ImageNetSubset-LT and Food101-LT with imbalance factor $\rho = 100$, memory budget $n_\varepsilon = 20$ evaluated under Learning From Scratch (\textit{LFS}) and Learning From Half (\textit{LFH}). \textbf{\red{Best}} and \textbf{\blue{Second Best}} results are marked. }
    \label{tab:results_all}
    \vspace{-2ex}
\end{table*}
\vspace{-0.3cm}
\section{Experiments}
\label{sec:experiments}
\vspace{-0.1cm}
\subsection{Experimental Setups}
\label{subsec: exp setup}
\vspace{-0.1cm}
\textbf{Evaluation protocol.} In this work, we adopt two widely used protocols for CIL including (1) learning from scratch (\textit{LFS})~\cite{ICARL} and (2) learning from half (\textit{LFH})~\cite{rebalancing, douillard2020podnet}. The \textit{LFS} equally split all the classes into $N$ tasks to incrementally train a model from scratch. The \textit{LFH} first trains the model with the initial half of the classes and then equally divides the remaining half of the classes into $N$ tasks. For both cases, the model is evaluated on all classes seen so far after learning each task without knowing the task identifier. We apply \textit{growing memory} with $n_\varepsilon$ exemplars per class selected by Herding~\cite{ICARL} for both \textit{LFS} and \textit{LFH}. The discussion about \textit{fixed memory} setup can be found in \textit{Appendix}.

\noindent \textbf{Datasets.} We evaluate our method on three public datasets including CIFAR100~\cite{CIFAR}, ImageNetSubset with 100 classes from ImageNet~\cite{IMAGENET1000} and Food101~\cite{Food-101}. Specifically, we first follow~\cite{liu2022long} to construct CIFAR100-LT, ImageNetSubset-LT and Food101-LT, which are the long-tailed versions of the original balanced datasets by removing training samples with an imbalance factor~\cite{cui2019class} $\rho = \frac{n_{max}}{n_{min}} = 100$ where $n_{max}$ and $n_{min}$ denote the maximum and minimum number of training data per class, respectively. For all three datasets, we use $N = \{10, 20\}$ tasks for \textit{LFS} and $N = \{5, 10\}$ tasks for \textit{LFH}. The test sets remain unchanged with the original class-balanced distributions. 

\noindent \textbf{Implementation details.}
We utilize ResNet-32~\cite{RESNET} for CIFAR100-LT and ResNet-18~\cite{RESNET} for ImageNetSubset-LT and Food101-LT. Both feature extractor and classifiers are trained from scratch using Equation~\ref{eq: z} in an end-to-end fashion. To ensure a fair comparison, we adopt the same class order and setups from~\cite{liu2022long} where we train 160 epochs for CIFAR100-LT with an initial learning rate of 0.1 and then reduced by a factor of 10 at the 80th and 120th epochs. For ImageNetSubset-LT and Food101-LT, we train 90 epochs with an initial learning rate of 0.1 which is reduced by 10 at the 30th and 60th epoch. A consistent batch size of 128 with an SGD optimizer is used across all experiments. For simplicity, we set our hyper-parameters with $\gamma =1$ and $\lambda_b =1$ for all experiments. Each experiments are conducted three times to report average performance. All results are obtained by reproducing the original methods under the same setting based on the framework in~\cite{survey_2020, liu2022long}. 

\subsection{Experimental Results}
\vspace{-0.1cm}
Table~\ref{tab:results_all} summarizes the  Average Accuracy (ACC)~\cite{GEM} on CIFAR100-LT, ImageNetSubset-LT, and Food101-LT with imbalance factor $\rho=100$, memory budget $n_\varepsilon = 20$ per class. We observe the \textit{LFS} is more challenging for CIL as the performance under \textit{LFH} protocol consistently higher than \textit{LFS}, underscoring the advantage of pre-training with a larger portion of classes in the initial incremental step. The 2-stage module~\cite{liu2022long} shows the effectiveness in addressing imbalanced data due to its additional learnable layer to mitigate bias. Notably, LUCIR-2stage~\cite{rebalancing}, which employs a cosine classifier with normalized weights, performs well in handling imbalanced data. Our method achieves promising results under \textit{LFH} and significant improvements under \textit{LFS} even without necessitating extra training stages and parameters. This demonstrates the potential of our end-to-end approach for more efficient training methodologies in CIL, especially as the task complexity scales up to learn an increasing number of classes from a scratch model. 

In Figure~\ref{fig:results_all}, we visualize the classification accuracy on all classes seen so far at each incremental step by varying the imbalance factor $\rho \in \{50, 150\}$ and the memory budget $n_\varepsilon \in \{10, 50\}$. While generally increasing the memory budget results in better performance, the improvements could be marginal for some methods such as SSIL~\cite{SSIL} and MAFDRC~\cite{MAFDRC}. This paradox is attributed to the exacerbation of exemplar set imbalance by using a larger memory budget $n_\varepsilon$ where the instance-rich classes retain more exemplars than instance-rare classes. On the other hand, the increase of imbalance factor $\rho$ results in a noticeable degradation in performance. Despite these challenges, our method is able to manage diverse learning environments posed by both memory budget and imbalance factors to consistently achieve the best performance at each incremental step. 

Additional results for (i) conventional CIL setup ($\rho = 1$), (ii) long-tail recognition, and (iii) computation and memory evaluations can be found in \textit{Appendix}.

\begin{figure}[t!]
\begin{center}
  \includegraphics[width=1.\linewidth]{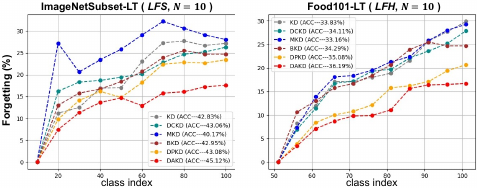}
  \vspace{-3ex}\caption{The Forgetting~\cite{GEM} (\%) at each incremental step by comparing our proposed DAKD with variants of distillation. The average classification accuracy (ACC) is shown in the legend ($\bullet$).}
  \label{fig:forgetting}
  \vspace{-0.4cm}
\end{center}
\end{figure}

\subsection{Ablation Study}
\label{subsec: ablation}
We evaluate each of our components including (i) the addition of our DAKD loss, (ii) the Decomposed Gradient Reweighting (DGR) to separately balance the gradients for new and learned tasks instead of treating all classes together (GR) for intra-phase imbalance, and (iii) the significance of maintaining the discrimination of gradients from knowledge distillation loss ($\triangle$(DAKD)) as described in Section~\ref{subsubsec: integrated gradient}. Specifically, we consider GR and original knowledge distillation (KD) as the baseline to evaluate the contributions by adding each component from our method. The results are summarized in Table~\ref{tab:ablation}, where we observe performance improvements contributed by each component. Notably, the GR exhibits better performance compared to DGR due to the potential underfitting of new classes which received insufficient emphasis during CIL. We also observe improved performance by maintaining the discriminativeness of the gradient from DAKD to capture and leverage information from the original training distribution, thus effectively preserving learned knowledge. 

\noindent\textbf{Variants of knowledge distillation loss:} Additionally,  we assess the effectiveness of DAKD to mitigate forgetting under CIL by replacing it with existing variants of logit-based knowledge distillation loss including (i) the balanced knowledge distillation (BKD)~\cite{BKD}, (ii) multi-level logit distillation(MKD)~\cite{MKD}, (iii) decoupled knowledge distillation (DCKD)~\cite{DCKD}, and (iv) decomposed knowledge distillation(DPKD)~\cite{DKD}. The results are visualized in Figure~\ref{fig:forgetting} ($\rho = 100, n_\varepsilon = 20$) where we measure the forgetting rate~\cite{GEM} at each incremental learning phase. Remarkably, we notice that the MKD, DCKD, and BKD achieve comparable or even worse results compared to the original KD loss as they assume the student and teacher model observe the same training distribution, which does not hold under CIL. While DPKD marks some improvements, its effectiveness remains hampered by imbalanced data. Overall, our DAKD achieves the lowest forgetting and best ACC by factoring lost training data distribution. 

         

\begin{table}[t]
    \centering
    \scalebox{.73}{
    \begin{tabular}{ccccccc}
        \hline
        \multicolumn{3}{c}{$\rho=100 \quad n_\varepsilon = 20$}&\multicolumn{2}{c}{\textbf{Food101-LT}} & \multicolumn{2}{c}{\textbf{ImageNetSubset-LT}} \\
        & &  &\multicolumn{2}{c}{\textit{LFS}} &\multicolumn{2}{c}{\textit{LFH}} \\ 
        \cdashline{4-7}
         DAKD & DGR & $\triangle$(DAKD) &  $N=10$ & $N=20$ & $N=5$  & $N=10$  \\
         \hline
         \small{(KD)} & \small{(GR)} & & 27.77 & 24.62 & 47.42 & 45.17  \\
         \small{(KD)} & \checkmark &  & 28.45 & 25.47 & 48.64 & 46.76  \\
          \checkmark & \small{(GR)} &  & 28.02 & 26.08 & 48.72 & 46.53 \\
          \checkmark & \small{(GR)} & \checkmark & 28.96 & 27.14 & 49.59 & 47.37 \\
         \checkmark  &\checkmark &\checkmark & \textbf{29.05} & \textbf{26.42} & \textbf{50.57} & \textbf{49.13} \\
         
        \hline
    \end{tabular}
    }
    \vspace{-0.1cm}
    \caption{Ablation study on Food101-LT and ImageNetSubset-LT}
    \vspace{-0.3cm}
    \label{tab:ablation}
\end{table}

\begin{figure}[t!]
\begin{center}
  \includegraphics[width=1.\linewidth]{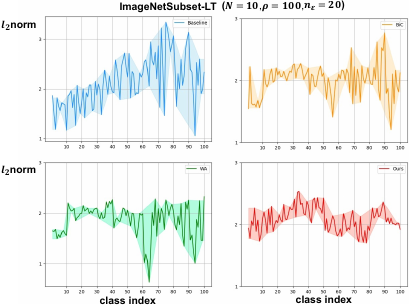}
  \vspace{-2ex}\caption{The $l_2$ norm of learned weight vectors after incrementally learning $N=10$ tasks on ImageNetSubset-LT under \textit{LFS} protocol. The shaded area shows the variations of each curve. }
  \label{fig:weight_norm}
      \vspace{-0.8cm}
\end{center}
\end{figure}

\noindent\textbf{Bias correction effects:} Finally, we examine the effectiveness of our method in rectifying the biased weights in the fully connected layers under CIL. Specifically, we compare with Baseline (fine-tuning), BiC~\cite{BiC} and WA~\cite{mainatining} to show the variance of $l_2$ norm for weight vectors corresponding to each class during CIL as shown in Figure~\ref{fig:weight_norm}. The Baseline method shows significant variation both within and between tasks. While BiC and WA address the variation between tasks through post-hoc bias correction, they do not effectively resolve the variation within individual tasks. Our method, in contrast, achieves more uniform weight vectors across both intra-task and inter-task learning, contributing to our best overall performance in imbalanced CIL.

%% file: sec/6_conclusion.tex
\vspace{-1ex}
\section{Conclusion}
\vspace{-1ex}
\label{sec:conclusion}

In this work, we study class-incremental learning (CIL) when applied to imbalanced data to address both intra-phase and inter-phase imbalances by reweighting the gradients in FC layer to foster a balanced optimization and learn unbiased classifiers. Additionally, we introduce a distribution-aware knowledge distillation loss that dynamically modulates the loss intensity in proportion to the extent of training data attrition to further mitigate imbalanced catastrophic forgetting. Our method shows consistent improvements under CIL and proves effective in long-tailed recognition. Overall, our findings underscore the importance of addressing data imbalance in CIL and pave the way for more robust and equitable class-incremental learning models.

\section*{Acknowledgement}
This project was supported by the National Institutes of Health (Grant No. U24CA268228). Special thanks are extended to Professor Fengqing Zhu for invaluable comments and advice on the manuscript, with a regrettable oversight of not being included in the author list.  